%% file: main_paper.tex
\documentclass[10pt,twocolumn,letterpaper]{article}

\usepackage{wacv}
\usepackage{times}
\usepackage{epsfig}
\usepackage{graphicx}
\usepackage{amsmath}
\usepackage{amssymb}
\usepackage{booktabs}
\usepackage{graphics} 
\usepackage{mathptmx} 
\usepackage{multirow}
\usepackage{makecell}
\usepackage{scalerel} 
\usepackage{enumitem} 

\def\wacvPaperID{570} 

\wacvalgorithmstrack   

\wacvfinalcopy 

\ifwacvfinal
\usepackage[breaklinks=true,bookmarks=false]{hyperref}
\else
\usepackage[pagebackref=true,breaklinks=true,colorlinks,bookmarks=false]{hyperref}
\fi

\begin{document}

\title{\LaTeX\ Author Guidelines for WACV Proceedings}




\author{Vladimir Somers\\
EPFL \& UCLouvain \& \\ Sportradar\\
{\tt\small vladimir.somers@epfl.ch}
\and
Christophe De Vleeschouwer\\
UCLouvain, Belgium\\
{\tt\small christophe.devleeschouwer} \\ {\tt\small @uclouvain.be}
\and
Alexandre Alahi\\
EPFL, Switzerland \\
{\tt\small alexandre.alahi@epfl.ch}
}

\title{\LARGE \bf
Body Part-Based Representation Learning for Occluded Person Re-Identification
}


\maketitle
\thispagestyle{empty}

\input{sections/placeholders}

                    
\input{sections/abstract}

\input{sections/introduction_new}
\input{sections/related_work}

\input{sections/methodology}

\input{sections/experiments}
\input{sections/conclusion}




{\small
\bibliographystyle{ieee}
\bibliography{references}
}

\input{sections/supplementary_materials}


\end{document}

%% file: sections/placeholders.tex

\newcommand\model{BPBreID}
\newcommand\G{G}
\newcommand\ffg{f_{fg}}
\newcommand\fgl{f_{gl}}
\newcommand\fcc{f_{cc}}
\newcommand\Y{Y}
\newcommand\scores{M} 
\newcommand\A{A}
\newcommand\F{F}
\newcommand\glb{hol}
\newcommand\lcl{part}
\newcommand\reid{ReID}


%% file: sections/abstract.tex
\begin{abstract}




Occluded person re-identification ({\reid}) is a person retrieval task which aims at matching occluded person images with holistic ones.
For addressing occluded {\reid}, part-based methods have been shown beneficial as they offer fine-grained information and are well suited to represent partially visible human bodies.
However, training a part-based model is a challenging task for two reasons. 
Firstly, individual body part appearance is not as discriminative as global appearance (two distinct IDs might have the same local appearance), this means standard \reid\ training objectives using identity labels are not adapted to local feature learning. 
Secondly, \reid\ datasets are not provided with human topographical annotations.
%
In this work, we propose {\model}, a body part-based {\reid} model for solving the above issues. 
We first design two modules for predicting body part attention maps and producing body part-based features of the \reid\ target.  
We then propose GiLt, a novel training scheme for learning part-based representations that is robust to occlusions and non-discriminative local appearance.
Extensive experiments on popular holistic and occluded datasets show the effectiveness of our proposed method, which outperforms state-of-the-art methods by $0.7\%$ mAP and $5.6\%$ rank-1 accuracy on the challenging Occluded-Duke dataset.
Our code is available at \url{https://github.com/VlSomers/bpbreid}.



\end{abstract}

%% file: sections/introduction_new.tex
\section{Introduction} \label{section:intro}


\begin{figure}[h!]
\begin{center}
\includegraphics[width=0.99\linewidth]{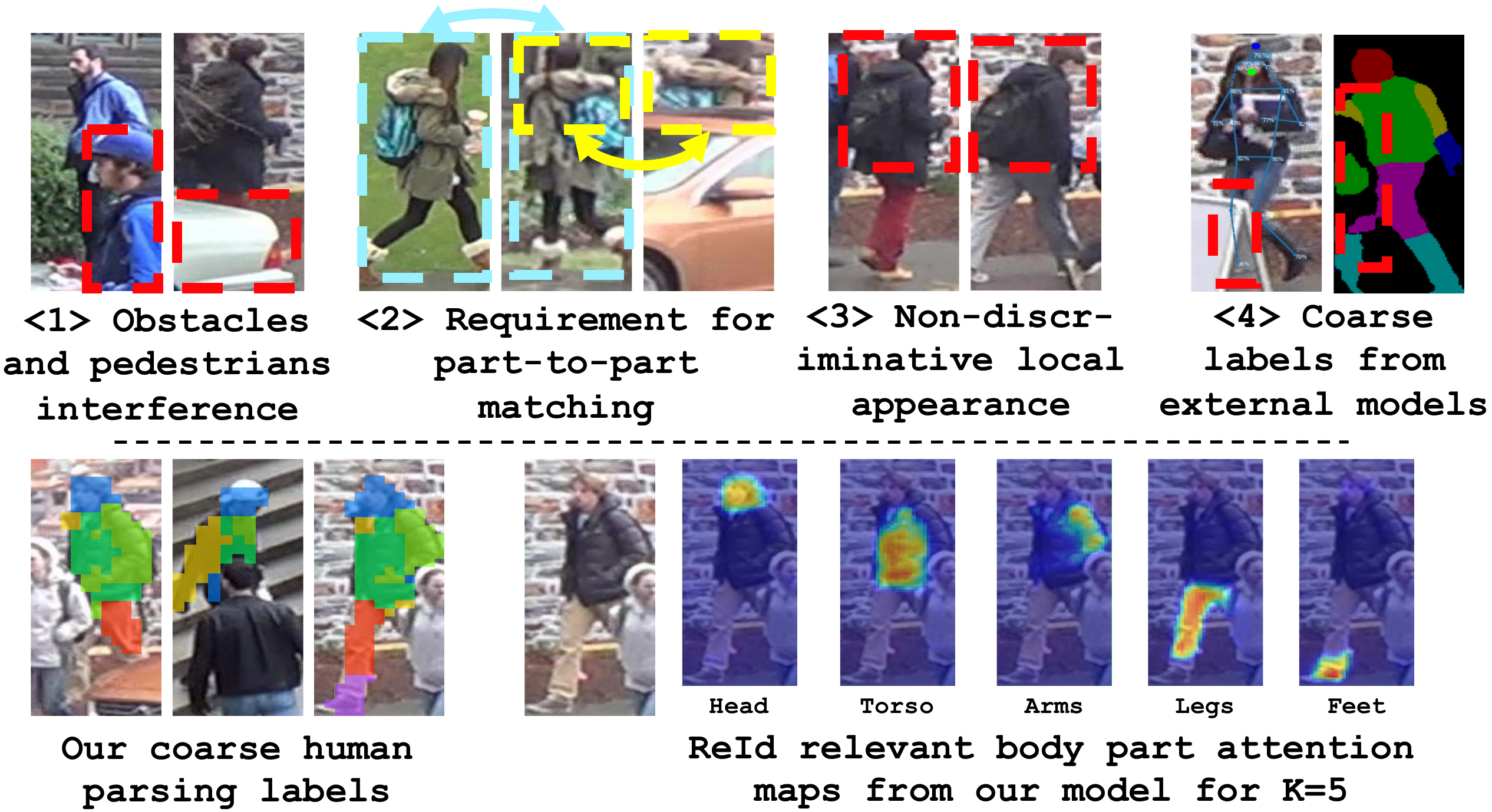}
\end{center}
  \caption{
  Overview of key concepts in our work.
  First row illustrates the four challenges of occluded and part-based {\reid} that our proposed method is trying to address. Second row illustrates our pre-generated human parsing labels and the ReID-relevant soft attention maps produced by our model {\model}.
  }
\label{fig:pull}
\end{figure}

Person re-identification \cite{Ye2021, occluded_reid_survey}, or \reid, is a person retrieval task which aims at matching an image of a person-of-interest, called the query, with other person images from a large database, called the gallery. 
\reid\ has important applications in smart cities for video-surveillance \cite{market1501, duke-mtmc} or sport understanding \cite{deepsportradarv1, soccernet22}.
Person re-identification is generally formulated as a representation learning task and is very challenging, because person images generally suffer from background clutter, inaccurate bounding boxes, pose variations, luminosity changes, poor image quality, and occlusions \cite{occluded_reid_survey} from street objects or other people. 


For solving the \reid\ task, most methods adopt a global approach \cite{BoT, triplet}, learning a global representation of the target person as a single feature vector.
However, these methods are unable to address the challenges caused by occlusions for two reasons, both depicted in Figure \ref{fig:pull}:

$\langle1\rangle$ \textit{Obstacle and pedestrians interference}: the globally learned representation might include misleading appearance information from occluding objects and pedestrians.

$\langle2\rangle$ \textit{Requirement for part-to-part matching}: 
When comparing two occluded samples, it is only relevant to compare body parts that are visible in both images.
Global method cannot achieve such part-to-part matching, because the same global feature is used for every comparison.


To deal with the above issues, part-based approaches \cite{PCB, ISP, HLGAT}, have shown promising results.
These part-based methods address the \reid\ task by producing multiple local feature vectors, i.e., one for each part of the input sample.
However, learning such part-based representations involves dealing with two crucial challenges:

$\langle3\rangle$ \textit{Non-discriminative local appearance}: 
Standard {\reid} losses, such as the id or triplet losses, work with the assumption that different identities have different appearance, and consequently that their corresponding global feature vectors are different.
However, this assumption is broken when working with part-based feature vectors, because two persons with different identities might have very similar appearance on some of their body parts, as depicted in Figure \ref{fig:pull}.
Because local appearance is not necessarily discriminative, standard {\reid} losses used for learning global representations do not scale well to local representation learning.
The specificity associated to learning local features and its impact on the choice of the training loss has been overlooked in previous part-based {\reid} works and we are the first to point it out.
To address these issues, we propose \textit{GiLt}, a novel training loss for part-based methods.
GiLt is designed to be robust to occlusions and non-discriminative local appearance, and is meant learn a set of local features that are each representative of their corresponding local parts, while being discriminative when considered jointly.

$\langle4\rangle$ \textit{Absence of human topology annotation}:
Part-based method generally rely on spatial attention maps to perform local pooling within a global feature map and build body part features of the {\reid} target.
However, no \reid\ dataset is provided with annotations regarding the local region to pool, and generating such annotation with external pose information or part segmentation tools yields inaccurate results due to the domain variation and poor image quality.
Moreover, body part-based feature pooling fundamentally differ from pixel-accurate human parsing.
Indeed, the spatial attention maps have to localize the body part in the image, but also to identify the feature vectors that best represent discriminant characteristics of the body part appearance.
Therefore, an ideal attention map is not necessarily an accurate segmentation shape.
Previous ReID works exploiting human parsing to build part-based features have either (i) used directly the output of a pose estimation model as local attention masks, without adapting it to handle the ReID task \cite{PVPM, HOReID, PGFA}, or (ii) learned local features with part discovery, without human topology prior \cite{PAT, ISP, DLPAR}.
In this work, we propose a body part attention module trained with a novel dual supervision, using both identity and coarse human parsing labels.
This module demonstrates how external human semantic information can be effectively leveraged to produce {\reid}-relevant body part attention maps.



Finally, we combine this body part attention module and GiLt (\underline{G}lobal-\underline{i}dentity \underline{L}ocal-\underline{t}riplet  ) loss to build our Body Bart-Based {\reid} model called {\model}, which effectively addresses all four challenges introduced before.
%
%
We summarize the main contributions of our work as follows:
\setlist{nolistsep}
\begin{enumerate}[noitemsep]

\item For the ReID task, we are the first to propose a soft attention trained from a dual supervision, to leverage both identity and prior human topology information. Our work demonstrates that this approach outperforms all previous part-based methods.
\item We propose a novel \textit{GiLt} strategy for training part-based method. GiLt is robust to occlusions and non-discriminative local appearance and is straightforward to integrate with any part-based framework.
\item {\model} outperforms state-of-the-art methods by $0.7\%$ mAP and $5.6\%$ rank-1 on the Occluded-Duke dataset. 
Our {\model} codebase has been released to encourage further research on part-based methods.
\end{enumerate}

%% file: sections/related_work.tex
\section{Related Work}
\paragraph{Part-based feature alignment in {\reid}:}
To solve the spatial misalignment issue, several works \cite{PCB, DAReID, PGFA, Su2017, Zhao2017, Zheng2017, DSA-reID, HOReID, PGFA, SRNet, PVPM} adopt fixed attention mechanisms, using pre-determined pixel partitions of the input image and applying part pooling for generating local feature representations.
These methods achieve poor feature selection and alignment, because the resulting attention maps are not meant for pooling {\reid}-relevant body part features. 
%
%
To solve those issues, other works \cite{Suh2018, Zhu2020, ISP, PFD} use attention mechanisms trained in an end-to-end fashion for generating attention maps that are specialized towards solving the \reid\ task.
Some of these approaches \cite{Suh2018, Zhu2020, PFD} include a pose estimation backbone as a parallel branch, which is jointly trained with the appearance backbone branch in an end-to-end fashion on the \reid\ dataset. 
However, the parallel branch induces a significant computational overhead and these methods do not address the occluded \reid\ problem explicitly.
%
Other part-based methods learn local features via part discovery in a self-supervised way, without human topology prior \cite{ISP, PAT, DLPAR}.
Such approach might introduce alignment errors, missed parts and background clutter.
%
Different from previous works, our part-based features are built by an attention branch which (i) is trained explicitly to pool local features that are relevant for the {\reid} task, and (ii) leverages external human parsing labels to bias the spatial attention in focusing on prior body regions.



\vspace{-10pt}
\paragraph{Local feature learning in {\reid}:} 
%
Identity loss and batch-hard triplet loss \cite{triplet} are two popular objectives for training \reid\ models that are also applied to part-based methods \cite{EXAM, PGFA, SRNet, HOReID, ISP, HPNet, MHSA-Net, PVPM} for learning local representations.
%
%
Most of these methods \cite{PCB, SGAM, PGFA, EXAM} solely apply an identity loss on each part-based features. 
As a consequence, they are more sensitive to non-discriminative body parts and miss out the benefits \cite{triplet, BoT} of the triplet loss as a complementary deep metric learning objective for {\reid}.
%
To deal with incomplete information of part features, some works \cite{ISP, MHSA-Net} propose to apply triplet and identity losses on a combined embedding, resulting from concatenation or summation of local features.
A specialized Improved Hard Triplet Loss (IHTL) is proposed in \cite{MHSA-Net} for training part-based feature, but this objective cannot cope well with occluded or similar samples.
Finally, \cite{HPNet} applies both triplet and identity losses on combined part features, but does not use holistic features during training, which renders their training scheme less robust to inaccurate body part predictions and heavy occlusions.
Our proposed GiLt training procedure aim at solving above issues and addressing the lack of consensus regarding the choice of losses to adopt for training part-based methods.
Finally, it is worth noting that other works \cite{PAT, HOReID} take an opposite approach to deal with standard {\reid} losses being unsuitable for non-discriminative body parts appearance.
They solve this by constraining each part-based feature to be discriminative on its own, by either having each of them attending simultaneously to multiple body regions \cite{PAT} or by adding high-order information in each local feature via message-passing \cite{HOReID}.

%% file: sections/methodology.tex
\section{Methodology} \label{methodology}



The overall architecture of our model {\model} is depicted in Figure \ref{fig:model_trained_attention}. 
It comprises two modules: the body part attention module described in Section \ref{section:attention_module} and the global-local representation learning module described in Section \ref{section:rep_learning}.
The overall training procedure of {\model} is described in Section \ref{section:training} and the procedure used at inference for computing query to gallery distance is described in Section \ref{section:testing}.

\subsection{Body Part Attention Module} \label{section:attention_module} 

The body part attention module takes as an input the feature map extracted by the backbone and outputs a set of attention maps highlighting the body parts of the \reid\ target.
This module consists of a \textit{pixel-wise part classifier} trained with a \textit{body part attention loss} using our coarse \textit{human parsing labels}.
We detail these three components below.
Because our model is trained end-to-end, the body part attention module also receives a training signal from the {\reid} loss, which uses the identity labels, as described in Section \ref{section:training}.
This attention branch is therefore trained from a \textbf{dual supervision}, with both a body part prediction objective and a \reid\ objective.
As a result of the dual supervision, this module generates attention maps that are more relevant to the \reid\ task than the attention maps we would obtain using the fixed output of a pre-trained human parsing model.
This module is depicted in the top left part of Figure \ref{fig:model_trained_attention}.

\subsubsection{Pixel-wise Part Classifier} \label{section:parts_classifier}
The body part attention module takes in input the appearance map $\G$, which is a tensor $R^{H \times W \times C}$ produced by a feature extractor.
For each pixel $(w, h)$ in the appearance map $\G$, a pixel-wise part classifier predicts if it belongs to the background or to one of the $K$ body parts, which means there are $K+1$ target classes, with the class at index 0 being the background.
A 1x1 convolution layer with parameters $P \in R^{(K+1) \times C}$ followed by a $softmax$ is applied on $\G$ to obtain the classification scores $\scores \in R^{H \times W \times (K+1)}$:
{\small{
\begin{equation} \label{eq:cls_scores}
    \scores = \text{softmax}(\G P^{T})\ .
\end{equation}
}}
%
These $K+1$ probability maps $\scores_k$ indicates therefore which pixels belong to which body parts (or to the background).

\subsubsection{Human Parsing Labels} \label{section:labels}
Human parsing labels $Y$ $\in R^{H \times W}$, required for training our part attention module, are generated with the PifPaf \cite{pifpaf} pose estimation model, following a process detailed in the supplementary materials.
$Y(h, w)$ is set to \{1, ..., K\} if spatial location $(h, w)$ belong to one of the $K$ body parts or $0$ for background.
Human semantic regions are defined manually for a given value of $K$.
For instance, with $K=8$, we define the following semantic regions: \{\textit{head}, \textit{left/right arm}, \textit{torso}, \textit{left/right leg} and \textit{left/right feet\}}.
These coarse human semantic parsing labels are illustrated in Figure \ref{fig:pull} for $K=5$.

\begin{figure*}[t!]
\begin{center}
\includegraphics[width=0.9\linewidth]{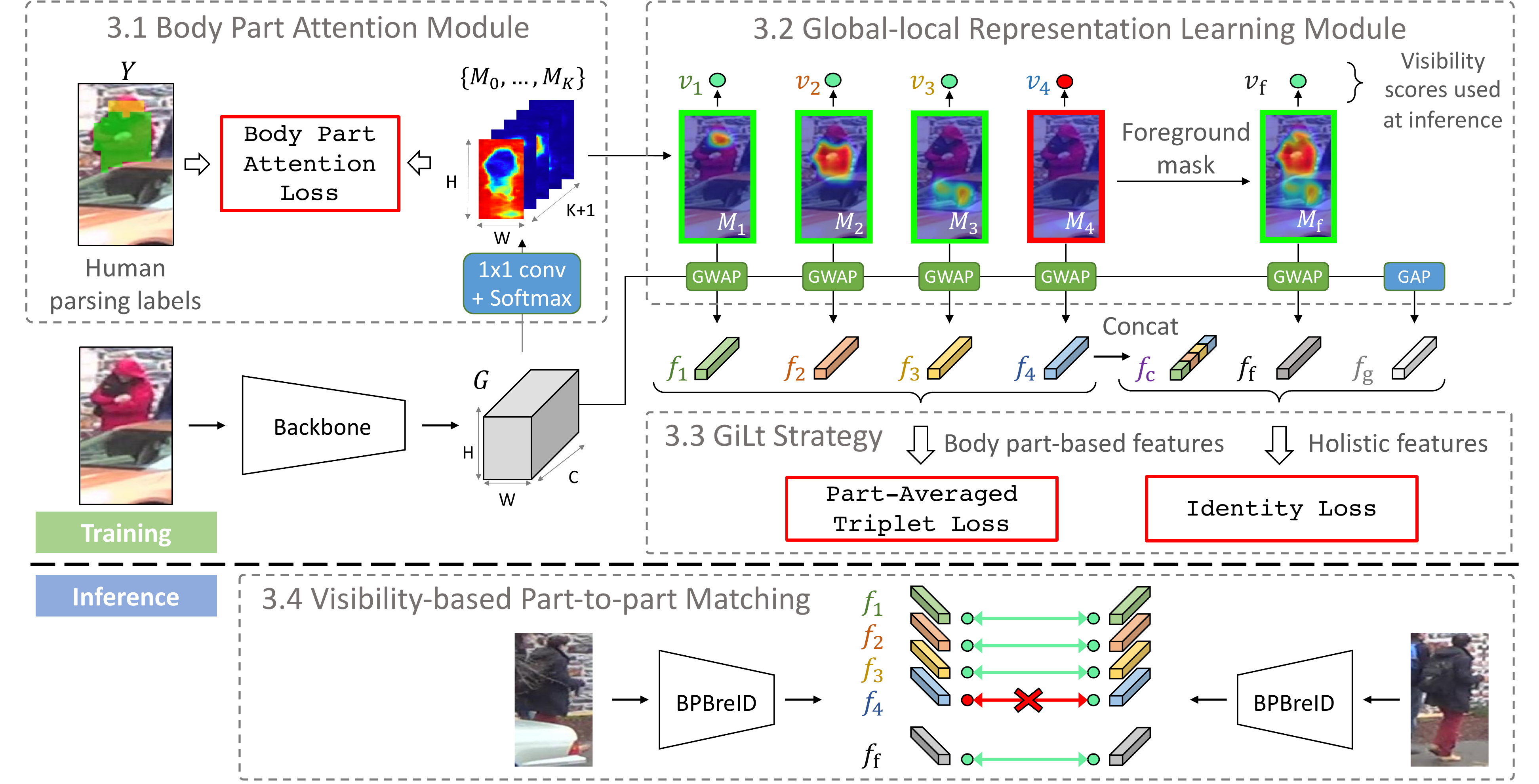}
\end{center}
  \caption{Structure of \model\, with detailed architecture and training procedure in the top part, and inference procedure in bottom part.
  The model consists of a \textit{body part attention module} for body part attention maps and a \textit{global-local representation learning module} for producing holistic features \{$f_{\text{g}}$, $f_{\text{f}}$, $f_{\text{c}}$\} and body part-based features \{$f_1$, ..., $f_K$\} together with their visibility scores \{$v_{\text{f}}$, $v_1$, ..., $v_K$\}.
  For holistic features, "$\text{g}$" stands for "global", "$\text{f}$" for "foreground" and "$\text{c}$" for "concatenated".
  \textit{GWAP} stands for global weighted average pooling.
  The network is trained in an end-to-end fashion using a \textit{body part attention loss} for supervising part prediction, a standard \textit{identity loss} on holistic features and a \textit{part-averaged triplet loss} on body part-based features. 
  Query to gallery distance is computed at inference using a \textit{part-to-part matching strategy} for comparing only mutually visible body parts. 
  Green/red color depict visible/invisible body parts.
  Each component of the architecture is framed with a grey rectangle, with its name and a number referencing the section describing it.
  For conciseness, {\model} is represented here with $K=4$: \{head, torso, legs, feet\}.}
\label{fig:model_trained_attention}
\end{figure*}

\subsubsection{Body Part Attention Loss} \label{section:part_attention_loss}
The pixel-wise part classifier is supervised with a body part attention loss $L_{pa}$, which is in practice a cross-entropy loss with label smoothing \cite{label_smoothing, Adaimi2019RethinkingConfidence}, as formulated here: 
{\small{
\begin{equation} \label{eq:pa_loss}
\setlength{\jot}{5pt}
\begin{split}
L_{pa} = - \sum_{k=0}^{K} \sum_{h=0}^{H-1} \sum_{w=0}^{W-1} q_{k} \cdot log(\scores_{k}(h, w)) \ , \\
\text{with } q_{k} =
\begin{cases}
\ 1 - \frac{N - 1}{N}\varepsilon & \text{\ if \ } Y(h, w) = k \\
\ \frac{\varepsilon}{N} & \text{\ otherwise , }
\end{cases}
\end{split}
\end{equation}
}}
where the human parsing labels map $Y$ is described in Section \ref{section:labels}, $N$ is the batch size, $\varepsilon$ is the label smoothing regularization rate and $\scores_{k}(h, w)$ is the prediction probability for part $k$ at spatial location $(w, h)$, as described in Eq. (\ref{eq:cls_scores}).

  
\subsection{Global-local Representation Learning Module} \label{section:rep_learning}
The global-local representation learning module takes as input the body part attention maps generated by the previous module, and outputs holistic and body part-based features of the \reid\ target, together with a visibility score for each part.
It can be visualized in the top right part of Figure \ref{fig:model_trained_attention}.
Part-based representations, combined with their visibility scores, is our solution for achieving part-to-part matching, and solving challenges $\langle1\rangle$ and $\langle2\rangle$ from Section \ref{section:intro}.

\subsubsection{Holistic and Body Part-based Features} \label{section:features}
As described in Section \ref{section:parts_classifier}, the body part attention module produces $K$ spatial heatmaps highlighting the corresponding $K$ predicted body parts of the input image. 
We first combine the $K$ body part maps $\{\scores_1, ..., \scores_K\}$ in a single foreground heatmap $\scores_{\text{f}} \in R^{H \times W}$: $\scores_{\text{f}}(h, w) = max\big(\scores_1(h, w), ..., \scores_K(h, w)\big)$. 
%
%
%
These heatmaps are then used to perform $K+1$ \textit{global weighted average pooling} (denoted \textit{GWAP} in Figure \ref{fig:model_trained_attention}) of the appearance feature map $\G$, to obtain the foreground embedding $f_\text{\text{f}}$ and the $K$ body part-based embeddings \{$f_1$, ..., $f_K$\}:
{\small{
\begin{equation} \label{eq:f_i}
    f_i = \frac
    {\sum_{h=0}^{H-1} \sum_{w=0}^{W-1} \G(h, w) M_i(h, w)}
    {\sum_{h=0}^{H-1} \sum_{w=0}^{W-1} M_i(h, w)}, \ \forall \ i \in \{\text{f}, 1, \ldots, K\}\ .
\end{equation}
}}

The initial global appearance feature map $\G$ is also globally average pooled (GAP) to obtain the global embedding $f_{\text{g}}$: $f_{\text{g}} = GAP(G).$ 
%
%
A last embedding $f_{\text{c}} \in R^{(C \cdot K)}$ is also produced by concatenating the $K$ body part-based features along the channel dimension: $f_{\text{c}} = concat(f_1, ..., f_K)$.
%
%
%
Our global-local representation learning module produces therefore three holistic embeddings \{$f_{\text{g}}$, $f_{\text{f}}$, $f_{\text{c}}$\} and $K$ body part-based embeddings \{$f_1$, ..., $f_K$\}.
%

\subsubsection{Body Part Visibility Estimation} \label{section:visibility}
%
%
To detect occluded body parts, we compute a binary visibility score $v_i$ for each embedding, with $0$/$1$ corresponding to invisible/visible parts respectively.
In our {\model} model, visibility scores are only used at inference.
%
%
For all holistic embeddings, visibility scores are set to one, i.e., $v_{\text{g}} = v_{\text{f}} = v_{\text{c}} = 1$.
For body part-based features, visibility score $v_i$ with $i$ $\in$ \{$1, ..., K$\} is set to 1 if at least one pixel in $M_i$ has a value above threshold $\lambda_{v}$, which is empirically set to $0.4$, as formulated below:
{\small{
\begin{equation} \label{eq:binary_visibility_score}
    v_i = 
    \begin{cases}
        1 & \quad \text{if \ $\underset{h, w}{\mathrm{max}}(M_i(h, w)) > \lambda_{v} $} \\
        0 & \quad \text{otherwise .}
    \end{cases}
\end{equation}
}}
%



\input{sections/methodology_training_new_wacv}



\subsection{Visibility-based Part-to-Part Matching} \label{section:testing}

Given a query sample $q$ and a gallery sample $g$, pairwise distance is computed at inference by a visibility-based part-to-part matching strategy using the foreground embedding and the body part-based embeddings:
{\small{
\begin{equation} \label{eq:test_dist}
dist_{total}^{qg} = 
\frac{  \sum\limits_{i \in \{\text{f}, 1, ..., K\}} \Big(v^{q}_{i} \cdot v^{g}_{i} \cdot dist_{eucl}(f_{i}^{q}, f_{i}^{g}) \Big) }
{  \sum\limits_{i \in \{\text{f}, 1, ..., K\}} \big(v^{q}_{i} \cdot v^{g}_{i} \big) }\ .
\end{equation}
}}
Visibility scores $v^{q|g}_{i}$ are used to ensure that only mutually visible body parts are compared.
If there's no mutually visible part between the two samples, their distance is set to infinity.
The strategy is illustrated in the bottom part of Figure \ref{fig:model_trained_attention}.
Global and concatenated embeddings are not used at inference because they may convey information from occluding objects and pedestrians.




%% file: sections/methodology_training_new_wacv.tex
\subsection{Overall Training Procedure} \label{section:training}
The overall objective function used to optimize the network during training stage is formulated as follows:
{\small{
\begin{equation} \label{overall_loss}
L = \lambda_{pa} L_{pa} + L_{GiLt}\ ,
\end{equation}
}}
where $L_{pa}$ is the body part attention loss supervised with human parsing labels (introduced in Section \ref{section:part_attention_loss}) and $L_{GiLt}$ is our \textit{GiLt} loss, supervised with identity labels. 
Parameter $\lambda_{pa}$ is used to control the overall part attention loss contribution and is empirically set to $0.35$.

\subsubsection{GiLt Loss} \label{section:gilt}
To supervise model training with the identity labels, our GiLt loss relies on two losses: the popular identity classification loss and a custom part-averaged triplet loss, which is a variant of the batch hard triplet loss \cite{triplet}.
However, we must carefully choose which loss to apply on each of the $K+3$ embeddings produced by our model.

First, unlike other popular part-based method \cite{PCB, PGFA, HOReID, PVPM, VGTri, PAT, DAReID, AAformer}, we do not apply the identity loss on part-based features because of occlusions and non-discriminative local appearance, as introduced in Section \ref{section:intro}.
Indeed, a part-based feature is not always discriminative enough to identify a person, which renders an identity prediction objective impossible to fulfil.
Consequently, adding an identity loss on such local representation would be destructive to performance.
However, similar to most state-of-the-art {\reid} methods, we still benefit from the identity loss supervision by applying it on the holistic features.

Secondly, we apply the triplet loss constraint on part-based features, via our custom \textit{part-averaged triplet loss} detailed in Section \ref{section:part_triplet_loss}.
At inference stage, distance between samples will be computed using these part-based features, and it therefore make sense to optimize their relative distances directly with a triplet loss constraint.
However, we argue the triplet constraint should not be enforced on holistic embeddings because of occlusions.
Indeed, two holistic embeddings of the same identity will have intrinsically different representations if at least one of the two is partially occluded, because each embedding will represent a different subset of the whole target body.
Therefore, pulling those two holistic features close together in the feature space with a triplet loss would be destructive to performance.

In summary, we claim the best training strategy for part-based methods is to apply (i) the identity loss constraint on holistic features only and (ii) the triplet loss constraint on part-based features only, via a our custom part-averaged triplet loss.
We call this strategy \textit{Global-identity Local-triplet} or simply \textit{GiLt}, and formulate it in our GiLt loss:
{\small{
\begin{equation} \label{eq:id_loss}
L_{GiLt} = L_{id} + L_{tri} = \sum\limits_{i \in \{\text{g}, \text{f}, \text{c}\}} L_{CE}(f_i)\ + L_{tri}^{parts}({f_1, ..., f_{K}}) ,
\end{equation}
}}
%
where $L_{CE}$ is the cross-entropy loss with label smoothing \cite{label_smoothing} and BNNeck trick \cite{BoT}, and $L_{tri}^{parts}$ is our part-averaged triplet loss detailed further below.
$L_{id}$ optimizes the network to predict the input sample identity from each holistic embedding \{$f_{\text{g}}$, $f_{\text{f}}$, $f_{\text{c}}$\}.

%



We provide extensive ablation studies in Section \ref{section:ablation_studies} for validating our claim.
These experiments also demonstrate the superiority of our GiLt strategy for training part-based methods compared to other combination of triplet and identity losses.
To our knowledge, we are the first to suggest such combination of triplet and identity losses for training part-based methods.
We are also the first to conduct extensive experiments to demonstrate the impact of both losses on training performance when enforced on holistic and part-based embeddings.
GiLt is illustrated in Figure \ref{fig:model_trained_attention}.

%
\subsubsection{Part-Averaged Triplet Loss} \label{section:part_triplet_loss}
Our part-averaged triplet loss differ from the standard batch hard triplet loss \cite{triplet} w.r.t. the strategy used to compute the distance between two samples.
Indeed, it relies on the average of pairwise parts distances between two samples $i$ and $j$. 
This part-averaged distance is computed using all body part-based features \{$f_1$, ..., $f_K$\} jointly:
{\small{
\begin{equation} \label{eq:parts_dist}
d_{parts}^{ij} = \frac{\sum_{k=1}^{K} dist_{eucl}(f_{k}^{i}, f_{k}^{j})}{K}\ ,
\end{equation}
}}
where $dist_{eucl}$ refers to the euclidean distance.
Similar to \cite{triplet}, the part-averaged triplet loss is then computed using the hardest positive and hardest negative part-averaged distances $d_{parts}^{ap}$ and $d_{parts}^{an}$ respectively:
{\small{
\begin{equation} \label{eq:parts_triplet}
L_{tri}^{parts}(f^a_0, ..., f^a_{K}) = [d^{ap}_{parts} - d^{an}_{parts} + \alpha]_+\ ,
\end{equation}
}}
where the distances from anchor sample to the hardest positive and negative samples are denoted by $d^{ap}$ and $d^{an}$ respectively, and $\alpha$ is the triplet loss margin.
Therefore, our part-averaged triplet loss globally optimize an average of local distances between corresponding parts, and not a distinct triplet for each part, as adopted in \cite{HOReID, PAT} and shown to be inferior in Table \ref{table:components}, under "\textit{BPBreID w/o part-averaged triplet loss}".
This critical design choice gives each training step the opportunity to focus on the parts with most robust and discriminant features, which in turns mitigates the impact of occluded and non-discriminative local features.


%% file: sections/experiments.tex
\section{Experiments} \label{section:experiments}

%

                                        
\subsection{Datasets and Evaluation Metrics} \label{section:datasets}
We evaluate our model on the holistic datasets Market-1501 \cite{market1501} and DukeMTMC-reID \cite{duke-mtmc}, and the occluded datasets Occluded-Duke \cite{PGFA}, Occluded-ReID\footnote{Occluded-ReID has no train set, so we use Market-1501 for training.} \cite{Zhuo2018} and P-DukeMTMC \cite{Zhuo2018}.
We report two standards \reid\ metrics: the cumulative matching characteristics (CMC) at Rank-1 and the mean average precision (mAP). 
Performances are evaluated without re-ranking \cite{re-rank} in a single query setting.

\subsection{Implementation Details} \label{section:implementation}
%
%
\textbf{Model architecture}
A \textit{ResNet-50} (RN) \cite{resnet} is employed as the main backbone feature extractor.
The final fully connected layer and global average pooling layer are removed and the stride of the last convolutional layer is set to 1 instead of 2.
For a fair comparison with methods using heavier architecture or training, we also employ the  \textit{ResNet-50-ibn} (\textit{RI}) \cite{ibn} as in \cite{fastreid, LDS, OPReID}, and the \textit{HRNet-W32} (HR) \cite{hrnet} backbone as in \cite{ISP}. HRNet feature maps with higher resolution are particularly beneficial to {\model} for building fine-grained attention maps.
All backbones are pre-trained on ImageNet \cite{imagenet}.
The number of body parts $K$ is set to $5$ for holistic datasets and $8$ for occluded datasets.
An ablation study on $K$ is provided in the supplementary materials.


\textbf{Training procedure}
The training procedure is mainly adopted from BoT \cite{BoT}.
All images are resized to 256$\times$128 for \textit{ResNet-50} (\textit{RN}) and 384$\times$128 with \textit{HRNet-W32} (\textit{HR}) and \textit{ResNet-50-ibn} (\textit{RI}). 
Images are first augmented with random cropping and 10 pixels padding, and then with random erasing \cite{random-erasing} at $0.5$ probability. 
A training batch consists of 64 samples from 16 identities with 4 images each.
The model is trained in an end-to-end fashion for 120 epochs with the Adam optimizer on one NVIDIA Quadro RTX8000 GPU. 
The learning rate is increased linearly from $3.5\times10^{-5}$ to $3.5\times10^{-4}$ after 10 epochs and is decayed to $3.5\times10^{-5}$ and $3.5\times10^{-6}$ at $40^{th}$ epoch and $70^{th}$ epoch respectively.
The label smoothing regularization rate $\varepsilon$ is set to $0.1$ and triplet loss margin $\alpha$ is set to $0.3$.

\input{sections/sota_comparison}

\subsection{Ablation Studies} \label{section:ablation_studies}
In this section, we conduct some ablations studies using the Occluded-Duke dataset and ${\model}_{RN}$, to analyze the impact of our architectural choices on \reid\ performance.


\input{tables/test_emb_with_components}

\subsubsection{Components of {\model}} \label{section:learnable_vs_fixed_attention}
Performance gain related to different components of our model are reported in Table \ref{table:components}.
We adopt Bag of Tricks (BoT) \cite{BoT} as a baseline and build {\model} on top of it.


%
%

\textit{{\model} without learnable attention} is an alternative method where the $K$ body part probability maps $\{\scores_1, ..., \scores_K\}$ predicted by the body part attention module are replaced by fixed attention weights derived directly from PifPaf output, following a process detailed in the supplementary materials.
%
The decreased performance primarily reveals that the lack of end-to-end training on the attention weights leads to a discrepancy between the fixed attention masks and the ReID need in terms of backbone feature pooling.
This confirms that training the attention mechanism in an end-to-end fashion with both body part prediction and \reid\ as objectives leads to an attention mechanism which is more specialized towards solving the \reid\ task, with a better selection of discriminative appearance features.
%
%

\textit{{\model} without visibility scores} refers to a strategy where all embeddings are used at inference no matter their visibility, i.e., all visibility scores $v^{q|g}_{i}$ in Eq. (\ref{eq:test_dist}) are set to 1.
As expected, using noisy embeddings corresponding to non-visible parts dramatically reduces performance.
This validates the effectiveness of our visibility-based part-to-part matching strategy for solving challenges $\langle1\rangle$ and $\langle2\rangle$. 
Our attempts to account for the visibility scores at training did not lead to performance improvement, but remains a promising path for future research. 
We speculate this happens because (1) GiLt is already robust to occlusion and (2) most training samples in Occluded-Duke are non occluded.

Finally, \textit{{\model} without part-averaged triplet loss} refers to a model where part-based embeddings are supervised individually with a classic triplet loss \cite{triplet} instead of our part-averaged triplet loss described in Eq. (\ref{eq:parts_triplet}).
The difference between these two approaches lies in their underlying objective: the part-averaged triplet loss optimizes a global distance between two samples, which is computed using the average of local distances, whereas the standard triplet loss optimizes all local pairwise distances individually.
The later approach gives reduced performance because it renders the training procedure more sensitive to occluded body-parts and non-discriminative local appearance.
This last ablation test demonstrates the robustness of our part-averaged triplet loss to occlusions and non-discriminative local features, and therefore to solve challenges $\langle1\rangle$ and $\langle3\rangle$.

\subsubsection{Validation of the GiLt Strategy} \label{section:gilt_experiments}
\input{tables/gilt_experiments}
In this section, we study the impact of different combinations of the identity and triplet losses on the $K+3$ embeddings produced by our model.
Results are reported in Table \ref{table:gilt_experiments}.
We also report performance for the popular model architecture PCB \cite{PCB}, which partition the input image in six horizontal stripes, to demonstrate the superiority of our training scheme with other part-based architecture.
The original PCB paper suggest a simple identity loss applied on part-based embeddings only: the corresponding sub-optimal performance is reported in the second table row.
%
%
The experiment on the first row correspond to our \textit{GiLt} strategy described in Section \ref{section:training}: holistic features are supervised only with an identity loss and part-based features are supervised with our part-averaged triplet loss.
%
As demonstrated by experiments 1 to 4, triplet and identity losses are complementary to each other and best performance is reached when using them together. 
However, naively applying both losses on all embeddings (experiment 2) is a sub-optimal solution.
We can draw two conclusions from experiments 5 to 8, which are small variations of our GiLt strategy regarding the identity loss. 
First, applying the identity loss on all three holistic embeddings leads to a more robust training scheme and to better performance.
This experiment validates our choice of computing a global and a concatenated embeddings for training, even though we don't use them at inference.
Second, experiment 5 validates our intuition that using an identity loss on part-based features is harmful to performance, since it renders the training procedure sensitive to occlusions and to non-discriminative local features.
Experiments 9 to 12 validate our \textit{GiLt} strategy of enforcing the triplet loss constraint on part-based embeddings only.
\subsubsection{Discriminative Ability of Output Embeddings} \label{section:test_embeddings}

\begin{figure}
\begin{center}
\includegraphics[width=0.99\linewidth]{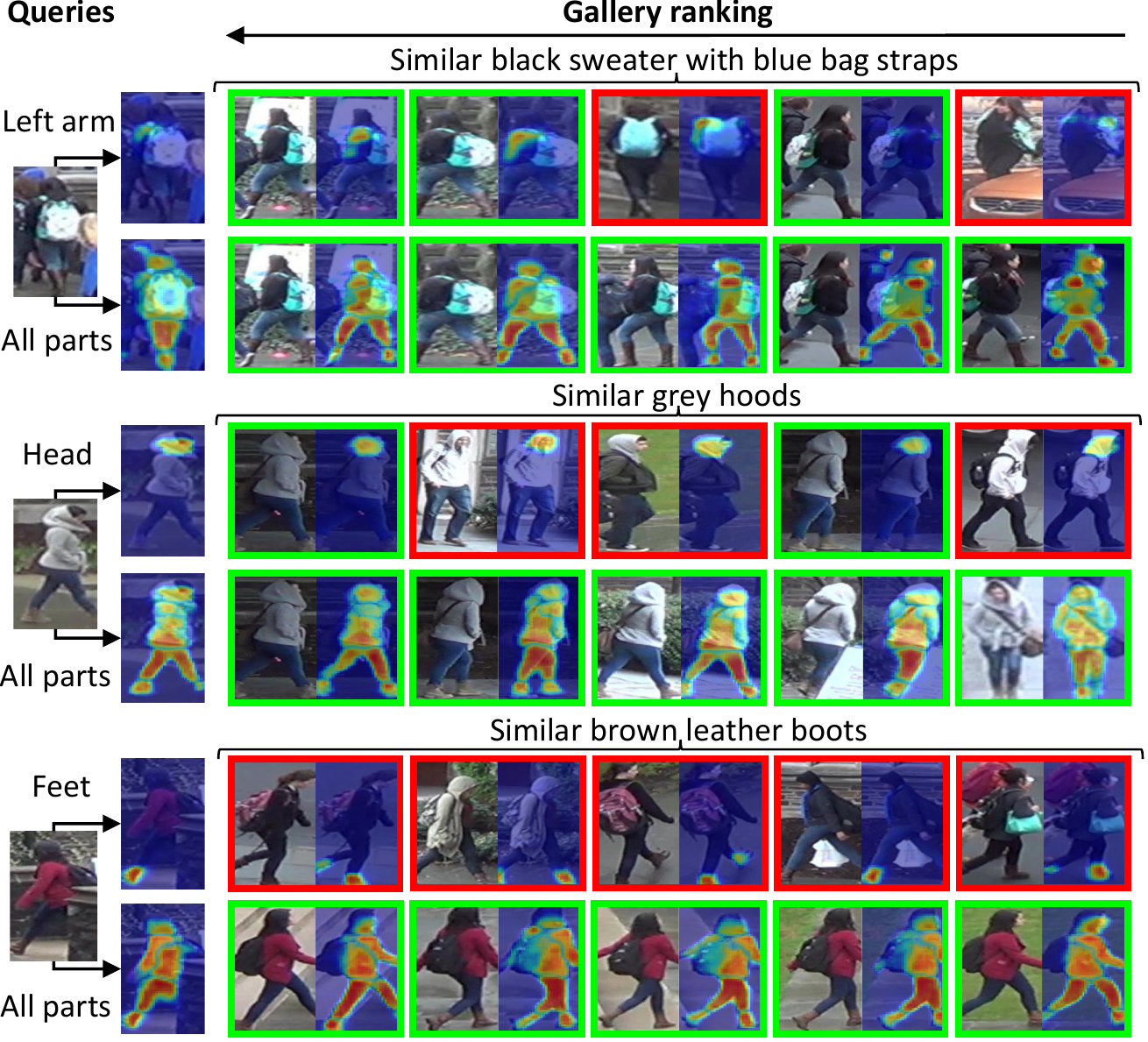}
\end{center}
  \caption{
    Visualization of ranking results based on individual body part-based embeddings (top row of each query) or all body part-based embeddings with the foreground embedding (bottom row of each query). 
  For the "all parts" rows, only the foreground attention map is displayed for conciseness. 
  In the top row of each query, the retrieved gallery samples are very similar w.r.t. the compared body part, but identities do not match because a single body part is not discriminative enough. 
  Green/red borders are correct/incorrect matches. Best viewed in color and zoomed in.
  }
\label{fig:body_parts_rankings}
\end{figure}

In this Section, we study the discriminative ability of the holistic and body part-based embeddings \{$f_{\text{g}}$, $f_{\text{f}}$, $f_{\text{c}}$, $f_1$, ..., $f_K$\} for $K=6$. 
For that purpose, we compute query-to-gallery samples distance using each embedding individually or combination of them, and report the corresponding ranking performance in Table \ref{table:test_time_embeddings}. 
When multiple embeddings are used, we compute the average distance weighted by visibility scores, as described in Eq. (\ref{eq:test_dist}).
As demonstrated in Table \ref{table:test_time_embeddings}, performance using holistic embeddings are sub-optimal because the global embedding is sensitive to background clutter, and the foreground embedding cannot achieve part-to-part matching.
The concatenated embedding fixes those issues but remains sensitive to occlusions because it contains noisy information from embeddings of non visible body parts.
As demonstrated in the table, using body part-based embeddings individually leads to sub-optimal performance.
Performance is better for embeddings from upper body parts, because (1) these parts are more discriminative and (2) lower body parts are very often occluded.
Using all parts embeddings \{$f_1$, ..., $f_6$\} leads to the best performance, because this strategy is the key to overcome the big challenges related to occluded re-id, i.e., (1) achieve feature alignment, (2) reduce background clutter and (3) compare only mutually visible body-parts.
Finally, adding information from the foreground embedding produces slightly better performance, because it helps in mitigating errors caused by failed body part prediction, and by image pairs having few or no mutually visible parts.
Figure \ref{fig:body_parts_rankings} illustrates some ranking results using these embeddings individually.

%% file: sections/sota_comparison.tex
\input{tables/sota_all_datasets_small}

\subsection{Comparison with State-of-the-Art Methods} \label{section:sota}

We compare our model in Table \ref{table:sota_all_datasets_small} with other {\reid} works and it ranks first overall.
Methods in the first part of the table use a ResNet-50 backbone with a training procedure similar to ours and BoT \cite{BoT}.
Methods in the second part either use arbitrary training procedures with bigger image size \cite{HPNet, PVPM, PGFL-KD, LDS}, more advanced backbones \cite{PFD, FED, ISP, OPReID}, or heavier architecture with additional backbones or branches \cite{PFD, PVPM, HOReID, LDS, PGFL-KD, PAT}.
\textbf{Occluded-Duke and P-DukeMTMC:}
Our model outperforms all previous part-based methods ($\ddagger$) on these two occluded datasets.
For methods using directly the output of a pose estimation model as local attention masks \cite{HOReID, PGFA, PVPM, PFD}, the lack of end-to-end training leads to sub-optimal attention maps in terms of {\reid}-relevant feature pooling. 
For methods producing local features via part discovery \cite{ISP, PAT}, not using prior human topology information renders their model more vulnerable to alignment errors, missed parts and background clutter.
Our work demonstrates that an end-to-end training of a spatial attention branch with both identity and human parsing labels is superior to previous architecture to perform {\reid}-relevant part-based pooling.
Recently, global methods ($\dagger$) designed specifically for the occluded {\reid} task \cite{HG, FED, LDS, OPReID} have shown promising performance compared to previous part-based methods. 
However, {\model} outperforms all of them as well, demonstrating the advantage of part-based methods to solve the occluded task, since global methods cannot achieve part-to-part matching.
%

\textbf{Market-1501 and DukeMTMC-ReID:} 
Our method outperforms all part-based methods ($\ddagger$) on both Market-1501 and DukeMTMC-ReID, except for PFD \cite{PFD} on Market-1501, which uses a much heavier architecture, with a ViT backbone and a HRNet-W48 parallel branch for pose estimation.
Regarding global methods ($\dagger$), {\model} outperform all of them on DukeMTMC-ReID, and achieves competitive performance on Market-1501, although the performance difference between most SOTA methods remains insignificant on it.
This demonstrates that part-based methods are a competitive choice for holistic person {\reid}.
%

\textbf{Occluded-ReID:}
This occluded dataset requires strong domain adaption capacity, since it does not provide a training set, and the Market-1501 dataset that is generally used for pre-training does not contain occluded samples.
All well-performing methods on Occluded-reID rely on the information from an external pose estimation model at inference \cite{PVPM, HOReID, PGFL-KD, PFD} or on occlusion data augmentation techniques \cite{FED, OPReID} to achieve robust part pooling on the new occluded domain.
Different from these methods, we don't use any external model at inference, nor occlusion data augmentation, but still achieve competitive performance.

%% file: tables/sota_all_datasets_small.tex
\begin{table}
\begin{center}
\caption{Comparison of {\model} with SOTA methods.
Symbols $\dagger$ / $\ddagger$ denote \textit{global} / \textit{part-based} methods respectively.
First, second and third best performance are indicated with $\textcolor{red}{^{1, 2, 3}}$ respectively.
}
\label{table:sota_all_datasets_small}
\setlength{\tabcolsep}{0.06em} 
{\fontsize{8.0}{10.3}\selectfont{
\begin{tabular}{|c|cc|cc|cc|cc|cc|}
\hline

\multicolumn{1}{|c|}{\multirow{4}{*}{\makecell[c]{Methods}}} &

\multicolumn{4}{c|}{\multirow{1}{*}{\makecell[c]{Holistic datasets}}} &
\multicolumn{6}{c|}{\multirow{1}{*}{\makecell[c]{Occluded datasets}}} \\
\cline{2-11}
& 
\multicolumn{2}{c|}{\multirow{2}{*}{\makecell[c]{Market\\-1501}}} &
\multicolumn{2}{c|}{\multirow{2}{*}{\makecell[c]{DukeMT\\-MC-ReID}}} &
\multicolumn{2}{c|}{\multirow{2}{*}{\makecell[c]{Occluded\\-Duke}}} &
\multicolumn{2}{c|}{\multirow{2}{*}{\makecell[c]{Occluded\\-reID}}} &
\multicolumn{2}{c|}{\multirow{2}{*}{\makecell[c]{P-Duke\\-MTMC}}}
\\
 & & & & & & & & & & \\
 \cline{2-11}
 & \multicolumn{1}{c}{R-1}&{mAP}&{R-1}&{mAP}&{R-1}&{mAP}&{R-1}&{mAP}&{R-1}&{mAP} \\
\hline
\hline
\multicolumn{11}{|c|}{\multirow{1}{*}{\makecell[c]{BoT-based training schemes with single ResNet-50 backbone}}} \\
\hline
 BoT \cite{BoT} $\dagger$ & 94.5 & 85.9 & 86.4 & 76.4 & 51.4 & 44.7 & 58.4 & 52.3 & 87.0 & 74.9 \\ 
\hline
SGAM \cite{SGAM} $\ddagger$ & 91.4 & 67.3 & 83.5 & 67.3 & 55.1 & 35.3 & - & - & - & - \\
PGFA \cite{PGFA} $\ddagger$ & 91.2 & 76.8 & 82.6 & 65.5 & 51.4 & 37.3 & - & - & 44.2 & 23.1 \\ 
MHSA \cite{MHSA-Net} $\ddagger$ & 94.6 & 84.0 & 87.3 & 73.1 & 59.7 & 44.8 & - & - & 70.7 & 41.1 \\ 
VGTri \cite{VGTri} $\ddagger$ & - & - & - & - & 62.2 & 46.3 & \textbf{81.0} & \textbf{71.0} & - & - \\
OAMN \cite{OAMN} $\dagger$ & 93.2 & 79.8 & 86.3 & 72.6 & 62.6 & 46.1 & - & - & - & - \\
HG \cite{HG} $\dagger$ & \textbf{95.6} & 86.1 & 87.1 & 77.5 & 61.4 & 50.5 & - & - & - & - \\ 
\hline
{\model}\textsubscript{RN} $\ddagger$ & 95.1 & \textbf{87.0} & \textbf{89.6} & \textbf{78.3} & \textbf{66.7} & \textbf{54.1} & 76.9 & 68.6 & \textbf{91.0} & \textbf{77.8} \\ 
\hline
\hline
\multicolumn{11}{|c|}{\multirow{1}{*}{\makecell[c]{Arbitrary backbones/training schemes or heavier architectures}}} \\
\hline
PVPM \cite{PVPM} $\ddagger$ & - & - & - & - & - & - & 66.8 & 59.5 & 85.1 & 69.9 \\
HOReID \cite{HOReID} $\ddagger$ & 94.2 & 84.9 & 86.9 & 75.6 & 55.1 & 43.8 & 80.3 & 70.2 & - & - \\
ISP \cite{ISP} $\ddagger$ & 95.3 & 88.6 & 89.6 & 80.0 & 62.8 & 52.3 & - & - & - & - \\
PAT \cite{PAT} $\ddagger$ & 95.4 & 88.0 & 88.8 & 78.2 & 64.5 & 53.6 & 81.6 & 72.1 & - & - \\ 
PGFL \cite{PGFL-KD} $\ddagger$ & 95.3 & 87.2 & 89.6 & 79.5 & 63.0 & 54.1 & 80.7 & 70.3 & 81.1 & 64.2 \\  
HPNet \cite{HPNet} $\ddagger$ & - & - & - & - & - & - & \textbf{87.3$\textcolor{red}{^{1}}$} & 77.4$\textcolor{red}{^{3}}$ & - & - \\ 
SSGR \cite{OPReID} $\dagger$ & \textbf{96.1$\textcolor{red}{^{1}}$} & 89.3 & 91.1 & 81.3 & 69.0 & 57.2 & 78.5 & 72.9 & - & - \\ 
FED \cite{FED} $\dagger$ & 95.0 & 86.3 & 89.4 & 78.0 & 68.1 & 56.4 & 86.3$\textcolor{red}{^{2}}$ &
 79.3$\textcolor{red}{^{2}}$ & - & - \\ 
LDS \cite{LDS} $\dagger$ & 95.8$\textcolor{red}{^{2}}$ & \textbf{90.3$\textcolor{red}{^{1}}$} & 91.5$\textcolor{red}{^{3}}$ & 82.5$\textcolor{red}{^{3}}$ & 64.3 & 55.7 & - & - & 91.9$\textcolor{red}{^{2}}$ & 82.9$\textcolor{red}{^{2}}$ \\ 
PFD \cite{PFD} $\ddagger$ & 95.5 & 89.7$\textcolor{red}{^{2}}$ & 91.2 & 83.2$\textcolor{red}{^{2}}$ & 69.5$\textcolor{red}{^{3}}$ & 61.8$\textcolor{red}{^{2}}$ & 81.5 & \textbf{83.0$\textcolor{red}{^{1}}$} & - & - \\ 
\hline
{\model}\textsubscript{RI} $\ddagger$ & 95.7 & 88.4 & 91.7$\textcolor{red}{^{2}}$ & 81.3 & 71.3$\textcolor{red}{^{2}}$ & 57.5$\textcolor{red}{^{3}}$ & 77.0 & 70.9 & 91.3$\textcolor{red}{^{3}}$ & 79.2$\textcolor{red}{^{3}}$ \\ 
{\model}\textsubscript{HR} $\ddagger$ & 95.7$\textcolor{red}{^{3}}$ & 89.4$\textcolor{red}{^{3}}$ & \textbf{92.4$\textcolor{red}{^{1}}$} & \textbf{84.2$\textcolor{red}{^{1}}$} & \textbf{75.1$\textcolor{red}{^{1}}$} & \textbf{62.5$\textcolor{red}{^{1}}$} & 82.9$\textcolor{red}{^{3}}$ & 75.2$\textcolor{red}{^{4}}$ & \textbf{93.0$\textcolor{red}{^{1}}$} & \textbf{83.2$\textcolor{red}{^{1}}$} \\ 

\hline
\end{tabular}
}}
\end{center}
\end{table}

%

%% file: tables/test_emb_with_components.tex
\begin{table}
\parbox{.55\linewidth}{
\centering
\caption{Ablation study for the main components of {\model} on Occluded-Duke.
For the experiment "\textit{w/o part-avgd triplet loss}", we replace our part-averaged triplet loss introduced in Eq. (\ref{eq:parts_triplet}) by a distinct triplet loss for each part.}
\label{table:components}
\setlength{\tabcolsep}{0.15em} 
{\fontsize{8.5}{10}\selectfont{
\begin{tabular}{|l|cc|}
\hline
Methods & R-1 & mAP\\
\hline
BoT \cite{BoT} baseline & 51.4 & 44.7 \\ 
\hline
{\model} & \textbf{66.7} & \textbf{54.1} \\
- w/o learnable attention & 51.6 & 39.2 \\
- w/o visibility scores & 52.6 & 45.3 \\ 
- w/o part-avgd triplet loss & 64.8 & 51.7 \\ 
\hline
\end{tabular}

}} 
}
\hfill
\parbox{.4\linewidth}{
\centering
\caption{Performance comparison for body part and holistic embeddings.}
\label{table:test_time_embeddings}
\setlength{\tabcolsep}{0.15em} 
{\fontsize{8.5}{10}\selectfont{
\begin{tabular}{|l|cc|}
\hline
Methods & R-1 & mAP\\
\hline
$f_{\text{g}}$ & 60.2 & 47.5 \\
$f_{\text{f}}$ & 64.1 & 49.7 \\
$f_{\text{c}}$ & 64.4 & 50.3 \\
$f_1$ (head) & 47.1 & 24.7 \\
$f_2$ (torso) & 52.0 & 31.7 \\
$f_3$ (left arm) & 55.7 & 34.4 \\
$f_4$ (right arm) & 56.7 & 34.1 \\
$f_5$ (legs) & 22.3 & 13.3 \\
$f_6$ (feet) & 16.1 & 9.0 \\
\{$f_1$, ..., $f_6$\} & 65.9 & 52.2 \\
\{$f_{\text{f}}$, $f_1$, ..., $f_6$\} & \textbf{66.1} & \textbf{52.5} \\
\hline
\end{tabular}
}}
}
\end{table}

%% file: tables/gilt_experiments.tex
\begin{table}
\begin{center}
\caption{
Impact of \textit{identity loss} and \textit{triplet loss} on training performance when applied selectively on holistic embeddings (\textit{global} "$\text{g}$", \textit{foreground} "$\text{f}$" and \textit{concatenated} "$\text{c}$") and \textit{body part-based embeddings} ("$p_{1,..,K}$"). 
Triplet loss on $p_{1,..,K}$ refers to our part-averaged triplet loss described in Section \ref{section:part_triplet_loss}.
Triplet loss on other embeddings ($\text{g}$, $\text{f}$ and $\text{c}$) refers to a standard batch-hard triplet loss \cite{triplet}.
Identity loss on $p_{1,..,K}$ refers to a identity loss applied individually on each part-based embeddings.
We also report performance for the popular part-based \reid\ architecture PCB \cite{PCB}, which does not use a foreground embedding. 
}
\label{table:gilt_experiments}
\setlength{\tabcolsep}{0.27em} 
{\fontsize{8.5}{10}\selectfont{
\begin{tabular}{|c|cccc|cccc|cc|cc|}
\hline
\multicolumn{1}{|c|}{\multirow{3}{*}{Idx}} & 
\multicolumn{4}{c|}{\multirow{2}{*}{\makecell[c]{Identity loss}}} &
\multicolumn{4}{c|}{\multirow{2}{*}{\makecell[c]{Triplet loss}}} &
\multicolumn{2}{c|}{\multirow{2}{*}{\makecell[c]{{\model} \\ (ours) }}} &
\multicolumn{2}{c|}{\multirow{2}{*}{\makecell[c]{PCB \cite{PCB}}}} \\
&&&&&&&&&&&& \\
\cline{2-13}
& \multicolumn{1}{c}{$\text{g}$} & {$\text{f}$} & {$\text{c}$} & {$p_{\scaleto{1,..,K}{4pt}}$} & {$\text{g}$} & {$\text{f}$} & {$\text{c}$} & {$p_{\scaleto{1,..,K}{4pt}}$} &
{R-1} & {mAP} & {R-1} & {mAP} \\

\hline
\hline
GiLt & \checkmark & \checkmark & \checkmark & & & & & \checkmark & \textbf{66.7} & \textbf{54.1} 
& 54.6 & \textbf{46.3}  \\ 
\hline
PCB & & & & \checkmark & & & & & 57.2 & 43.2 
& 51.2 & 40.8 \\ 
1 & & & & \checkmark & \checkmark & \checkmark & \checkmark & & 52.9 & 43.2 
& 50.2 & 40.9  \\ 
2 & \checkmark & \checkmark & \checkmark & \checkmark & \checkmark & \checkmark & \checkmark & \checkmark & 59.5 & 48.2 
& 51.1 & 42.8  \\ 
3 & \checkmark & \checkmark & \checkmark & \checkmark & & & & & 61.5 & 49.5 
& 52.1 & 44.8  \\ 
4 & & & & & \checkmark & \checkmark & \checkmark & \checkmark & 53.9 & 41.9 
& 45.5 & 37.6  \\ 
\hline
5 & \checkmark & \checkmark & \checkmark & \checkmark & & & & \checkmark & 61.8 & 49.4 
& 51.0 & 43.5  \\ 
6 & \checkmark & \checkmark & & & & & & \checkmark & 65.5 & 51.4 
& 52.9 & 43.5  \\ 
7 & \checkmark & & \checkmark & & & & & \checkmark & 56.5 & 41.9 
& - & - \\ 
8 & & \checkmark & \checkmark & & & & & \checkmark & 64.0 & 52.9 
& \textbf{56.2} & 46.2  \\ 
\hline
9 & \checkmark & \checkmark & \checkmark & & & & \checkmark & \checkmark & 66.2 & 53.3 
& 55.9 & 46.1  \\ 
10 & \checkmark & \checkmark & \checkmark & & & \checkmark & & \checkmark & 63.6 & 52.2 
& - & - \\ 
11 & \checkmark & \checkmark & \checkmark & & \checkmark & & & \checkmark & 64.0 & 52.4 
& 54.8 & 45.9  \\ 
12 & \checkmark & \checkmark & \checkmark & & & & \checkmark & & 65.3 & 52.9 
& 54.4 & 45.7  \\ 
\hline
\end{tabular}
}}
\end{center}
\end{table}

%% file: sections/conclusion.tex
\section{Conclusions}
In this work, we propose our model BPBreID to address the occluded person {\reid} task by learning body part representations and make two contributions.
First, we design a body part attention module trained from a dual supervision with both identity and human parsing labels.
With this attention mechanism, we show how external human semantic information can be effectively leveraged to produce {\reid}-relevant part-based features.
Second, we investigate the influence of triplet and identity losses for learning part-based features and provide a simple yet effective GiLt strategy for training any part-based method.
Our model achieves state-of-the-art performance on five popular {\reid} datasets.

\textbf{Acknowledgment:}
This work has been funded by Sportradar, by the Walloon region project ReconnAIssance, and by the Fonds de la RechercheScientifique – FNRS.

%% file: sections/supplementary_materials.tex
\clearpage
\appendix
\pagebreak
\section*{Supplementary materials}

Our code is available at \url{https://github.com/VlSomers/bpbreid} and is based on the \textbf{Torchreid}\footnote{https://github.com/KaiyangZhou/deep-person-reid} framework.
The clean and modular architecture of our framework with SOTA performance will hopefully attract researchers looking for a strong baseline to conduct further research on human-part based ReID.
In the next section, we provide further details on the generation of our human parsing labels. 
We also provide further experiments on the number of body parts defined by the hyper-parameter $K$, and qualitative assessment for the ranking performance and the attention maps.

\subsection*{Human Parsing Labels Generation with PifPaf}
Human parsing labels $Y$, required for training our part attention module, are generated using the 17 part confidence and 19 part affinity fields produced by the PifPaf \cite{pifpaf} pose estimation model.
These 36 part confidence and affinity fields are probability maps highlighting different human body region, i.e., 17 human keypoints and 19 joints between these keypoints.
For further details about the encoder part of the PifPaf model, we refer readers to \cite{pifpaf}.
We split these 36 heatmaps into $K$ groups and perform a pixel-wise max operation within each group to obtain $K$ new maps highlighting $K$ body regions.
These K maps are then concatenated to produce a tensor $E$ $\in$ $R^{H \times W \times K}$.
Each of the $K$ groups correspond to a human semantic region (i.e. body part).
These groups are defined manually for a given value of $K$.
Choosing $K$ and defining the right human semantic regions is therefore part of the model hyperparameter tuning process.
For instance, with $K=8$, we define the following semantic regions: \{head, left/right arm, torso, left/right leg and left/right feet\}.
Each element $(h, w, c)$ in $E$ indicates to which degree the spatial location (h, w) belongs to body part $c$.
We perform a final $argmax$ operation on $E$ to produce the human parsing label map:
{\small{
\begin{equation} \label{eq:labels}
    Y(h, w) = 
    \begin{cases}
    0 & \text{if } \underset{c}{\mathrm{max}}(E(h, w, c)) < \lambda_{t} \\
    1 + \underset{c}{\mathrm{argmax}}(E(h, w, c)) & \text{otherwise ,}
    \end{cases}
\end{equation}
}}
where pixels with none of the K channel values above a threshold $\lambda_{t}=0.5$ are considered background. 
An illustration of these coarse human semantic parsing labels is given in Figure \ref{fig:pull} for $K=5$.
If multiple persons are detected within a sample, we assume the \reid\ target is the pedestrian with its head closer to the top center part of the bounding box and remove labels from other persons.
We refer readers to our GitHub for more details about the human parsing labels generation strategy.

Instead of using PifPaf, we also tried using some popular human parsing models (\textbf{Densepose}\footnote{https://github.com/facebookresearch/DensePose} and \textbf{SCHP}\footnote{https://github.com/GoGoDuck912/Self-Correction-Human-Parsing}) to generate our human parsing labels, but obtained poor performance because of domain transfer and low image quality in the ReID datasets we target. Human parsing labels obtained with PifPaf gave the best results because it provides consistent predictions with few false negative on a wide range of image resolutions.

In experiment \textbf{"\textit{BPBreID without learnable attention}"} from Table \ref{table:components}, the $K$ body part probability maps $\{\scores_1, ..., \scores_K\}$ predicted by the body part attention module are replaced by the fixed tensor $E$ described above, on which a channel wise softmax is applied to produce fixed body part classification scores, used as attention weights.

\subsection*{Study on K, the number of body parts} \label{section:number_k}

In this Section, we study the impact of the number $K$ of body parts predicted by the body part attention module.
The body part attention module is trained using some pre-generated human parsing labels: different labels should therefore be used depending on the value $K$.
The human parsing labels are 2D human semantic segmentation maps, where each pixel is assigned an integer value from 0 to $K$, 0 being the background label and values from 1 to $K$ being labels for the K body regions.
These maps are therefore used to indicate to which body part each pixel of the input image belongs to.
Human parsing labels for a few samples are illustrated in Figure \ref{fig:pull}.
In Table \ref{table:number_k}, we report ranking performance on the Occluded-Duke dataset for various values of $K$ and the corresponding grouping strategy.
As demonstrated in this table, best performance is reached with $K=8$.
Other values of $K$ provide too low/high granularity and lead to reduced performance.

\begin{table*}[h!]
\begin{center}
\begin{tabular}{|l|cc|c|}
\hline
K & R-1 & mAP & Grouping strategy for defining human parsing training labels\\
\hline
2 & 58.3 & 49.0 & \makecell[l]{\{upper body (torso + arms + head), lower body (legs + feet)\}} \\ 
3 & 63.0 & 52.0 & \makecell[l]{\{head, middle body (torso + arms), lower body (legs + feet)\}} \\ 
4 & 64.3 & 52.9 & \makecell[l]{\{head, torso, arms, lower body (legs + feet)\}} \\ 
5 & 65.0 & 53.3 & \makecell[l]{\{head, torso, arms, legs, feet\}}\\ 
6 & 66.1 & 52.5 & \makecell[l]{\{head, torso, right arm, left arm, legs, feet\}}\\ 
8 & \textbf{66.7} & \textbf{54.1} & \makecell[l]{\{head, torso, right arm, left arm, right leg, left leg, right foot, left foot\}} \\ %
11 & 66.5 & 52.9 & \makecell[l]{\{head, upper torso, lower torso, upper right arm, lower right arm, \\ upper left arm, lower right arm, right leg, left leg, right foot, left foot\}}\\ 
\hline
\end{tabular}
\end{center}
\caption{Comparison on Occluded-Duke for different values of K, i.e., the number of body parts embeddings generated by our model, together with the grouping strategy used to generate the corresponding target human parsing labels. These labels are used to train the body part attention module and indicate to which human body region (or background) each pixel in the input image belongs to. The last column details the semantic meaning of each of the $K$ body parts.}
\label{table:number_k}
\end{table*}

\subsection*{Qualitative comparison of ranking performance} \label{section:ranking_performance}
\begin{figure*}[t!]
\begin{center}
\includegraphics[width=0.99\linewidth]{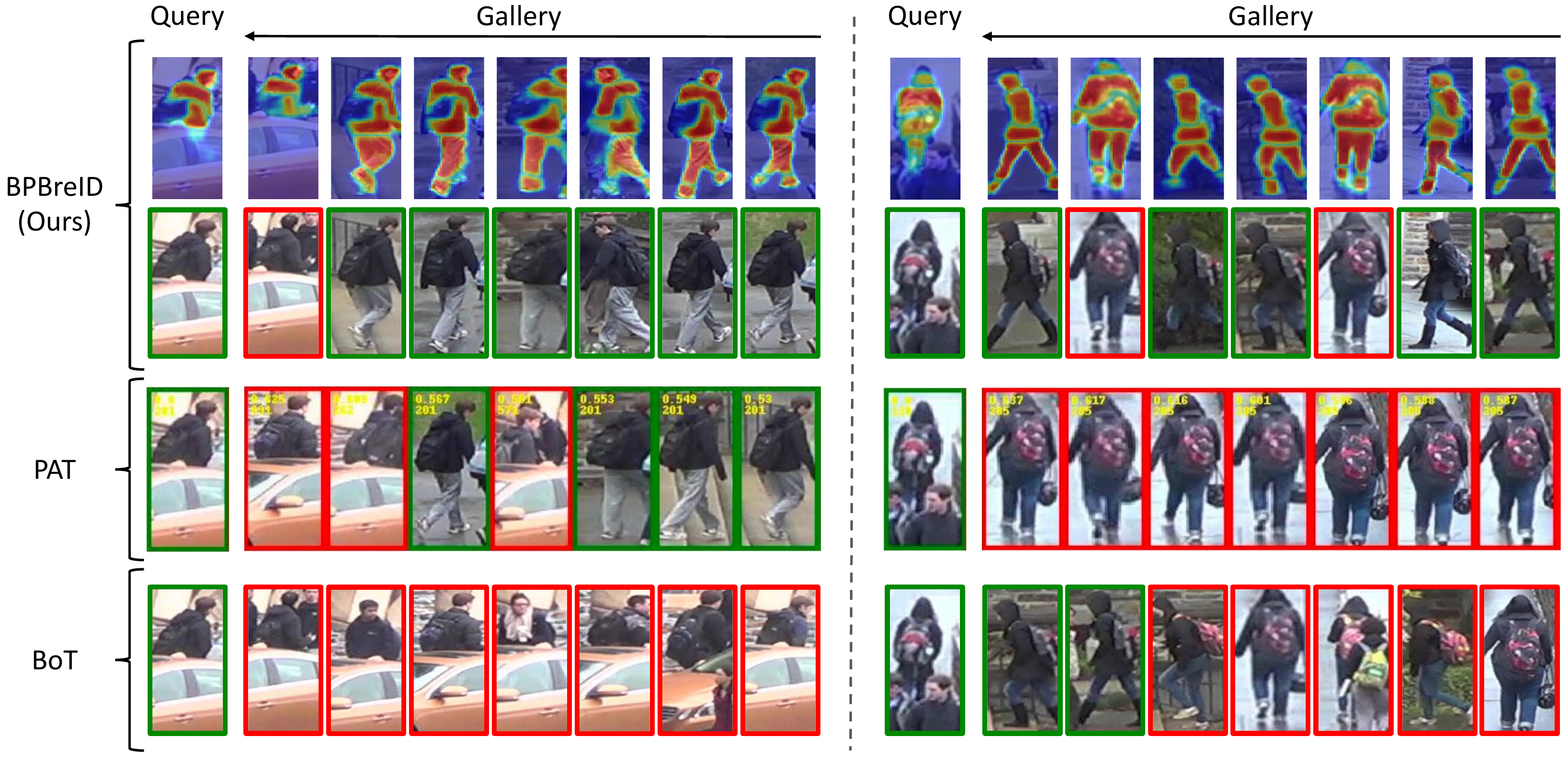}
\end{center}
   \caption{We compare the ranking performance of our model {\model} with other methods: the part-based transformer method with part discovery PAT \cite{PAT} and our baseline, the global method BoT \cite{BoT}. As illustrated in this figure, BoT cannot handle occlusions and PAT is inferior in terms of detecting and aligning fine-grained local appearance features.}
\label{fig:ranking_comparison}
\end{figure*}

We compare ranking performance of our model to other works in Figure \ref{fig:ranking_comparison}.

\subsection*{Qualitative comparison of attention maps} \label{section:ranking_performance}
\begin{figure*}[t!]
\begin{center}
\includegraphics[width=0.99\linewidth]{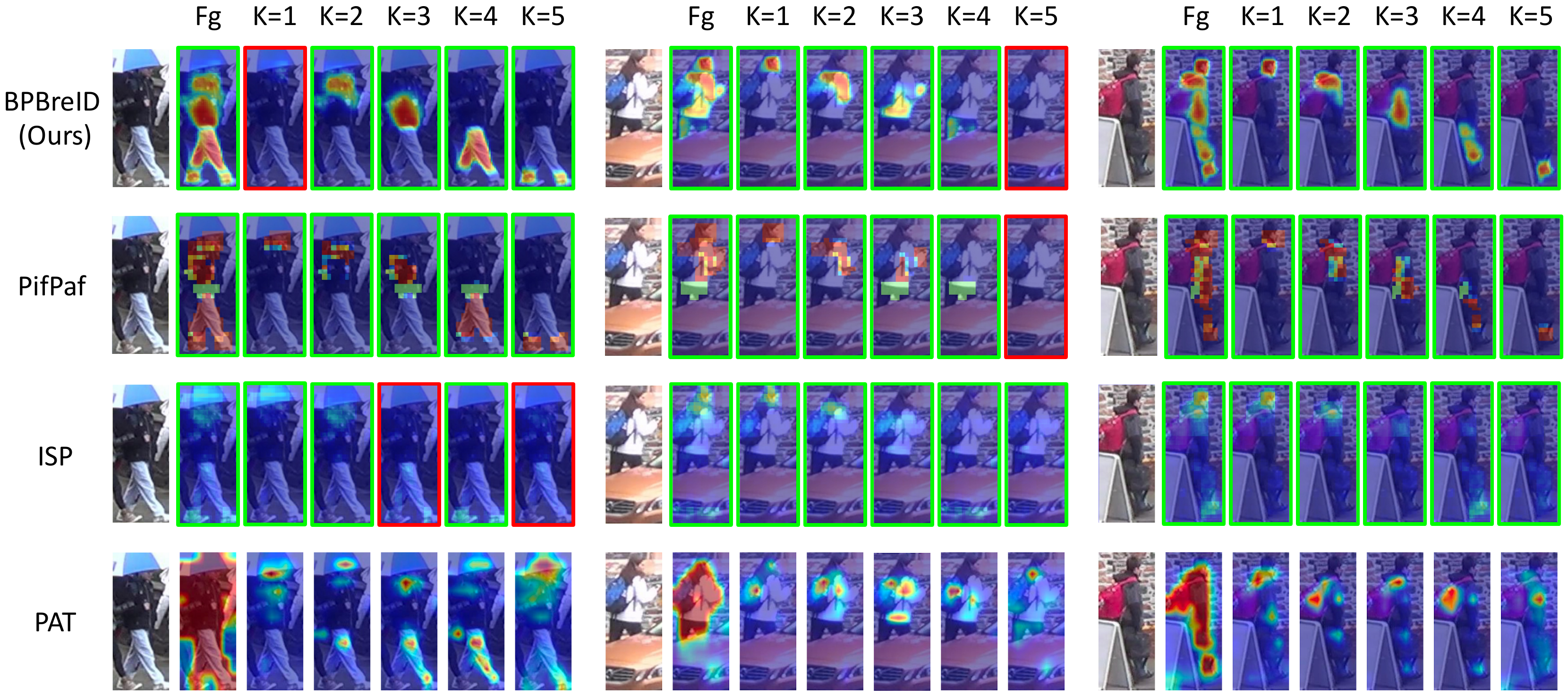}
\end{center}
   \caption{We compare the attentions maps produced by our model {\model} (on test images unseen at training) with the attention maps from other state-of-the-art part-based methods: ISP \cite{ISP} and PAT \cite{PAT}. "Fg" refers to the foreground attention maps, which is obtained by fusing maps from all parts together.
   Green/red borders illustrate visible/unvisible parts and no color is displayed for PAT because this method is not designed with a visibility score mechanism. Both ISP and PAT use part-discovery to define the human semantic regions, which can lead to missed part, background clutter or feature misalignment. As illustrated in this figure, our attention maps doesn't suffer from these issues. However, unlike these methods, our method only detects body parts and no belongings, such as bags or umbrellas. 
   Moreover, most part-based methods (such as PAT \cite{PAT}, ISP \cite{ISP}, HOReID \cite{HOReID}, ...) tries to make each part-based embedding discriminative on its own. 
   This is performed by either incorporating global information into each local embedding \cite{HOReID}, or by having each part attending to multiple regions of target person body \cite{PAT}, or by mining discriminative local features \cite{ISP}, as illustrated in this Figure.
   Different from these methods, we learn part-based embeddings that well represent their associated body-part, without the requirement of being discriminative on their own, but with the requirement of being discriminative when used as a whole.
   The PifPaf row illustrate the coarse PifPaf part confidence and affinity fields described in the first section of these supplementary materials (tensor $E$ for $K=5$), from which we derive our human parsing labels used at training. }
\label{fig:heatmaps}
\end{figure*}

We compare the attentions maps of our model to other works in Figure \ref{fig:heatmaps}.

